\documentclass[10pt,twocolumn,letterpaper]{article}

\usepackage{wacv}              

\makeatletter
\@namedef{ver@everyshi.sty}{}
\makeatother

\usepackage{times}
\usepackage{epsfig}
\usepackage{graphicx}
\usepackage{amsmath}
\usepackage{amssymb}
\usepackage[dvipsnames, svgnames, x11names]{xcolor}
\usepackage{booktabs}
\usepackage{enumerate}
\usepackage{enumitem}
\usepackage{makecell}
\usepackage{tabu}
\usepackage{multirow}
\usepackage{utfsym}
\usepackage{colortbl}
\usepackage{xcolor}
\usepackage{array}

\usepackage{url}
\usepackage{hyperref}
\hypersetup{
	final,
	breaklinks,
	colorlinks,
	linkcolor={Mahogany},
	citecolor={YellowGreen},
	urlcolor={CadetBlue}
}

\usepackage[capitalize]{cleveref}
\crefname{section}{Sec.}{Secs.}
\Crefname{section}{Section}{Sections}
\Crefname{table}{Table}{Tables}
\crefname{table}{Tab.}{Tabs.}

\begin{document}

\title{FreMIM: Fourier Transform Meets Masked Image Modeling \\
 for Medical Image Segmentation}

\author{
Wenxuan Wang\textsuperscript{1,\thanks{Equal Contribution.\textsuperscript{\dag}Corresponding author.}}
\quad Jing Wang\textsuperscript{1,*}
\quad Chen Chen\textsuperscript{2}
\quad Jianbo Jiao\textsuperscript{3}\\
\quad Yuanxiu Cai\textsuperscript{1} 
\quad Shanshan Song\textsuperscript{1}
\quad Jiangyun Li\textsuperscript{1\dag}\\
{\textsuperscript{1}School of Automation and Electrical Engineering, University of Science and Technology Beijing} \\
{\textsuperscript{2}Center for Research in Computer Vision, University of Central Florida} \\
{\textsuperscript{3}School of Computer Science, University of Birmingham} \\
\small{\texttt{s20200579@xs.ustb.edu.cn, chen.chen@crcv.ucf.edu, leejy@ustb.edu.cn}}
}

\maketitle

\begin{abstract}
    The research community has witnessed the powerful potential of self-supervised Masked Image Modeling (MIM), which enables the models capable of learning visual representation from unlabeled data. In this paper, to incorporate both the crucial global structural information and local details for dense prediction tasks, we alter the perspective to the frequency domain and present a new MIM-based framework named FreMIM for self-supervised pre-training to better accomplish medical image segmentation tasks. Based on the observations that the detailed structural information mainly lies in the high-frequency components and the high-level semantics are abundant in the low-frequency counterparts, we further incorporate multi-stage supervision to guide the representation learning during the pre-training phase. Extensive experiments on three benchmark datasets show the superior advantage of our FreMIM over previous state-of-the-art MIM methods. Compared with various baselines trained from scratch, our FreMIM could consistently bring considerable improvements to model performance. The code will be publicly available at \url{https://github.com/Rubics-Xuan/FreMIM}.
\end{abstract}


\section{Introduction}
\label{sec:intro}

Since Masked Language Modeling (MLM) obtained great success in the field of Natural Language Processing (NLP) \cite{devlin2018bert}, numerous works \cite{he2022masked, zhou2021ibot, peng2022beit, bao2021beit, xie2022simmim, chen2020generative} have transferred this idea to the vision domain, making Mask Image Modeling (MIM) an effective pre-training strategy. One of the most representative approaches is Masked Autoencoders (MAE) \cite{he2022masked}, which pre-trains the model by masking partial regions within an image and reconstructing them.
After the pre-training, the model is fine-tuned on various downstream tasks and achieves state-of-the-art (SOTA) performance. 
Following-up works mainly focus on improving the accuracy and efficiency by introducing new designs, such as ConvMAE~\cite{gao2022convmae} and Siamese Image Modeling~\cite{tao2022siamese}.

\begin{figure}[!tp]
    \centering
    \rotatebox{90}
    {
    \begin{tabu} to 0.6\linewidth{X[0.8c] X[1.2c]} 
        \scriptsize{(b) Ours} & \scriptsize{(a) MAE\cite{he2022masked}} \\ 
    \end{tabu}
    }
    \includegraphics[width=0.44\textwidth]{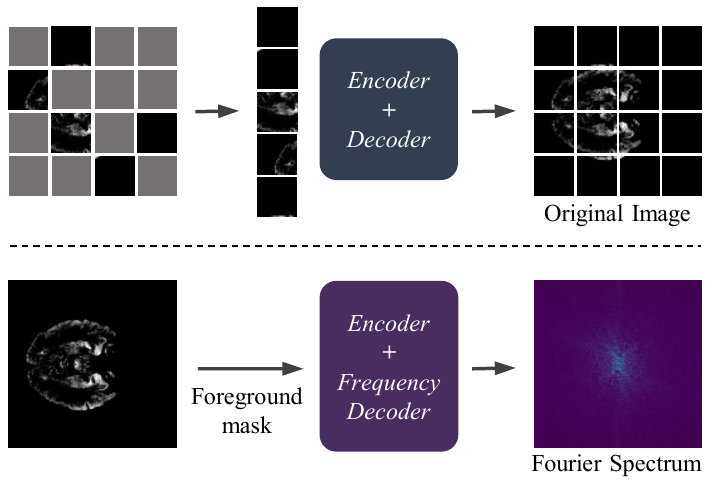}
    \vspace{-10pt}
    \caption{The comparison of key ideas between the MAE framework and our proposed FreMIM. (a) MAE: randomly masks the patch tokens and reconstructs raw pixels of the original image. 
    (b) Our FreMIM: randomly masks the \textbf{foreground pixels} and reconstructs the \textbf{Fourier spectrum} of the original image.}
    \label{fig1}
    \vspace{-16pt}
\end{figure}

Some recent works applied MAE-based methods for medical image analysis
\cite{zhou2022self,tang2022self,huang2022attentive} and achieved promising results across various benchmark datasets with different modalities, including computed tomography (CT)~\cite{nguyen2015segmentation} images, magnetic resonance imaging (MRI)~\cite{huo2017robust}, to name a few. 
Despite making methodological advancements and structural innovations, 
these methods have not essentially solved the key limitations of MAE. 
Although compared with other self-supervised learning (SSL) frameworks MAE can consistently help the model extract generally useful features even with few training samples (as proven by \cite{Kong2022UnderstandingMI}), to some extent, MAE solely takes raw pixels as reconstruction targets mainly depending on local feature representation rather than fully utilizing the global information.
Besides, since the model is expected to possess the ability to extract features with multiple semantic levels at different stages, only the output from the last stage is fed into the decoder for the reconstruction task, lacking the supervision from other stages to provide multi-scale information.
In summary, previous works \cite{he2022masked,bao2021beit,wei2022masked} crucially require a certain trade-off between the local details and contextual semantics, which leaves room for further improvement.
Furthermore, due to high acquisition costs and patients' privacy, the training samples of commonly small-scale medical datasets are relatively limited, but none of these previous works have taken this unique characteristic of medical image datasets into consideration and made tailored designs.

\begin{figure}[!tp]
\vspace{-5pt}
    \centering
    \rotatebox{90}
    {
    \begin{tabu} to 0.6\linewidth{X[0.8c] X[1.2c] } 
        \scriptsize{Spatial Domain} &  \scriptsize{Frequency Domain}  \\ 
    \end{tabu}
    }
    \includegraphics[width=0.43\textwidth]{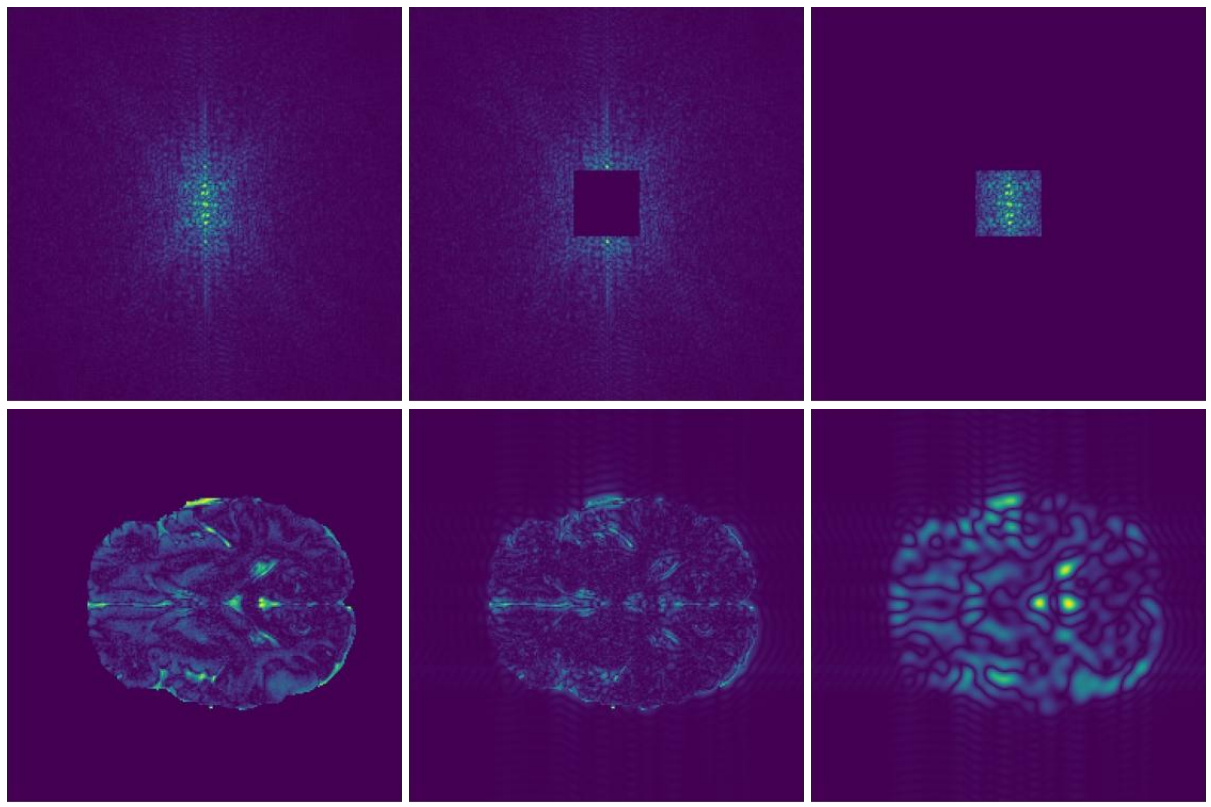}
    \begin{tabu} to 1.0\linewidth{X[2.0c] X[1.0c] X[1.6c]}
        \  \scriptsize{All frequency} &  \scriptsize{ high frequency} &  \scriptsize{low frequency}
    \end{tabu}
    \vspace{-15pt}
    \caption{The visualization of the whole Fourier spectrum, high-frequency components, and low-frequency counterparts respectively, the high/low-frequency components of which are acquired by applying the corresponding high/low-pass filters on the whole Fourier spectrum.
    As illustrated in the second row of the figure, the inspiration for our FreMIM comes from the observations that local details (like texture and contours) mainly lie in the high-frequency components while the global and smooth structural information is rich in the low-frequency counterparts.}
    \label{fig_method_highlow}
    \vspace{-10pt}
\end{figure}

Therefore, in order to fully exploit the potential of MIM-based methods for medical image segmentation under the circumstance of limited training samples, \textit{how to acquire the global information while preserving the detailed local features as much as possible} has become the key problem. 
Considering the nature of Fourier Transform in image processing, it might be a possible solution. 
As studied in lots of previous works \cite{si2022inception, chen2019drop, bullier2001integrated, bar2003cortical, kauffmann2014neural} and shown in Fig. \ref{fig_method_highlow}, the detailed texture information mainly lies in the high-frequency components and the low-frequency counterparts carry rich global information. 
Following this observation, an intuitive solution would be exploring the powerful potential of MIM coupled with Fourier Transform.

To this end, 
aiming at the joint modeling of both local and global features during SSL pre-training, 
we propose a new MIM-based framework conducted in the Fourier domain, namely \textit{FreMIM}, which to our knowledge is the first work to explore the potential of MIM with Fourier Transform for 2D medical image segmentation. 
Specifically, our FreMIM first masks out a portion of randomly selected image pixels and then predicts the corresponding missing frequency spectrum of the input image in the Fourier domain. 
Since medical images of the same organ essentially correspond to similar features, we conduct difficult cross-domain reconstruction tasks to avoid learning with shortcuts and achieve strong representation capability. 
Meanwhile, inspired by previous findings \cite{wang2020high} that the detailed structural information mainly lies in the high-frequency components and the high-level semantics are abundant in the low-frequency counterparts, the proposed bilateral aggregation decoder is leveraged to sequentially apply the Fourier Transform on the original image and employ low/high-pass filters on the converted Fourier spectrum to get the expected reconstruction target.
Such a multi-stage supervision approach could better guide the model pre-training, resulting in better representations for segmentation. Besides, we propose an effective foreground masking strategy as the alternative to the original random masking, which is proven to be more suitable for textures and details modeling for medical image segmentation. 
In summary, the main contributions of this work are summarized as follows:
\setlist{nolistsep}
\begin{itemize}[noitemsep,leftmargin=*] 

    \item We present the first study on exploring the powerful potential of masked image modeling with frequency domain for medical image segmentation tasks. The proposed FreMIM is a generic self-supervised pre-training framework that can be integrated with different model architectures (\ie both CNNs and Transformers).
	\item By designing a multi-stage supervision scheme coupled with a well-designed bilateral aggregation decoder, we propose a new cross-domain masking-reconstruction framework for masked image modeling paradigm.
	\item A simple yet effective masking strategy among foreground pixels is proposed as a better alternative to the original random masking pixels strategy, providing more precise and informative masks for the following self-supervised representation learning.
	\item Without introducing any extra training samples, extensive experiments on three benchmark datasets and three representative 2D baselines validate the effectiveness of the proposed FreMIM, outperforming other previous alternative self-supervised state-of-the-art approaches.
\end{itemize}

\section{Related Work}
\label{sec:formatting}

\subsection{Masked Image Modeling}
As a powerful self-supervised learning paradigm, 
MIM has attracted increasing community interest recently. By reconstructing the masked portion of images, models could learn informative feature representations that are favorable for various visual downstream tasks. 

\noindent \textbf{On Natural Images.}
Previous works of reconstruction targets could be divided into three categories, including discrete tokens\cite{bao2021beit,peng2022beit}, feature maps\cite{wei2022masked, zhou2021ibot}, and raw image pixels\cite{xie2022simmim, he2022masked}. For example, BEiT \cite{bao2021beit} and BEiTV2 \cite{peng2022beit} added a classifier to predict masked visual tokens, and it is supervised by the encoded image patches from offline tokenizer. Inspired by the self-distillation paradigm in DINO\cite{caron2021emerging}, iBOT \cite{zhou2021ibot} adopted a teacher-student framework to perform MIM. The teacher network serves as an online tokenizer to learn visual semantics from all image patches, while the student network only processes visible patches. Moreover, MaskFeat \cite{wei2022masked} first explored features as prediction targets. 
Besides, SimMIM \cite{xie2022simmim} discarded the tokenizer and patch classification, simply employing RGB values of raw pixels as predicted targets.
Without feeding masked tokens into the encoder, MAE \cite{he2022masked} designed a simple decoder to reconstruct image patches, leading to a considerable reduction of computation complexity during pre-training.

\noindent \textbf{On Medical Images.}
At the same time, various works \cite{haghighi2022dira, zhou2021preservational, zhou2023unified, zhou2022self, tang2022self, huang2022attentive} have explored the effectiveness of MIM pre-training on various medical benchmark datasets. 
Zhou et al.\cite{zhou2022self} applied MAE pre-training paradigm for medical image segmentation and significantly improved the results. 
Huang et al.\cite{huang2022attentive} proposed a manually settled attentive reconstruction loss that pays more attention to the informative regions.
Tang et al.\cite{tang2022self} explored the hierarchical structure for full extraction of image features and constructed a self-supervised pre-training framework with three proxy tasks. 
However, the random masking strategy of patches utilized previously is rough and may result in computation waste on the useless background. 
Considering that informative foreground and useless background are discriminated in medical images, we design a masking strategy among foreground pixels to obtain more effective masks, assisting models in better representation learning. 
Moreover, our method could cast off the reliance of the pre-training paradigm on specific model structures and consistently boost model performance, which is different from previous works (\eg Swin Transformer and CNN-based models can not be directly integrated with MAE).

\subsection{Fourier Transform}
Recently, a series of research\cite{rao2021global, zhou2022deep, jiang2021focal} have performed Fourier Transform on images and leveraged the frequency information to improve model performance and efficiency. For example, \cite {rao2021global} utilized Fast Fourier Transform (FFT) as the alternative to self-attention modules in the original Transformer, successfully acquiring global information with low computation costs.  
\cite{jiang2021focal} designed a novel focal frequency loss for Fourier spectrum supervision to improve popular image generative model performance. 

Inspired by these previous researches \cite{si2022inception, chen2019drop, bullier2001integrated, bar2003cortical, kauffmann2014neural}, we randomly mask the original image and reconstruct the Fourier spectrum in the frequency domain to help the model learn more generalized global representation in a cross-domain masking-reconstruction manner. In addition, multi-stage supervision coupled with leveraged specific characteristics of FFT (\ie high-pass and low-pass frequency components) is also proposed to better guide the model representation learning among different stages.

\section{Methodology}

\subsection{Preliminary: Fourier Transform}
\label{subsec:preliminaries}

Since Discrete Fourier Transform (DFT) plays a vital role in our proposed method, we first give a brief review of the 2D DFT that serves as an indispensable technique for traditional signal analysis. 
Given a 2D signal $\mathbf{F} \in \mathbb{R}^{W \times H}$, its corresponding 2D-DFT can be defined as:
\begin{equation}
	\small
	f(u, v)=\sum_{h=0}^{H-1} \sum_{w=0}^{W-1} F{(h, w)} e^{-j 2 \pi\left(\frac{uh}{H}+\frac{vw}{W}\right)},\label{eq:fft}
\end{equation}
where $F{(h, w)}$ represents the signal located at $(h, w)$ in $\mathbf{F}$, while the $u$ and $v$ are indices of horizontal and vertical spatial frequencies in the Fourier spectrum. 
Correspondingly, the 2D Inverse DFT~(2D-IDFT) is formulated as:  
\begin{equation}
	\small
	F(h, w)=\frac{1}{H W} \sum_{u=0}^{H-1} \sum_{v=0}^{W-1} f(u, v) e^{j 2 \pi\left(\frac{uh}{H}+\frac{vw}{W}\right)}.\label{eq:ifft}
\end{equation}

Both DFT and IDFT can be accelerated with their fast version, FFT algorithm~\cite{nussbaumer1981fast}. 
For medical images with various modalities, the Fourier Transform is operated on each channel independently. Besides, as already shown in previous works~\cite{si2022inception, chen2019drop, bullier2001integrated, bar2003cortical, kauffmann2014neural}, the detailed structural texture information of an image mainly lies in the high-frequency part of the Fourier spectrum while the global information is rich in the low-frequency counterpart. Fig.~\ref{fig_method_highlow} presents the visualization of this intriguing characteristic.


\subsection{The Proposed FreMIM}

\paragraph{Overall Architecture.}

\begin{figure*}[!tp]
    \centering
    \includegraphics[width=0.90\textwidth]{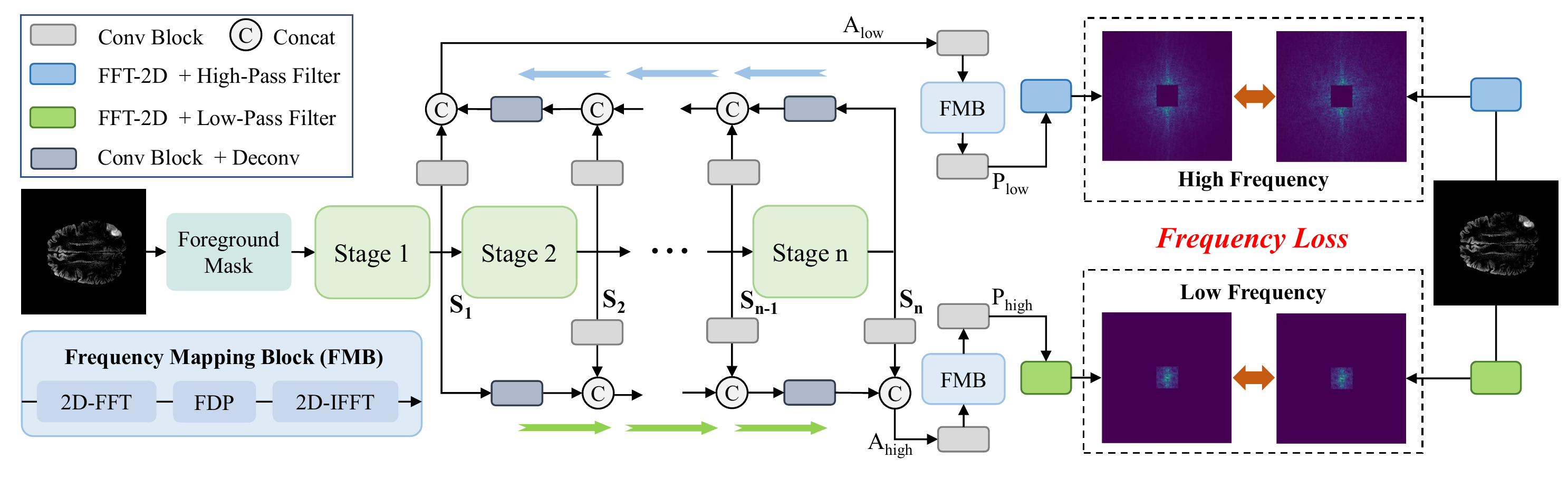}
    \vspace{-15pt}
    \caption{The overall architecture of our proposed FreMIM. 
    At first, the input medical image is corrupted by the foreground masking strategy and then fed into the encoder, which consists of several stages with a hierarchical structure. The captured feature maps at different stages (\ie $S_{1}$, $S_{2}$, ... $S_{n}$) are fused by a bilateral aggregation decoder to generate the aggregated high-level and low-level feature representations (\ie $A_{high}$ and $A_{low}$). For the fused feature of each semantic level, an FMB is applied respectively to learn its recessive information in the frequency domain, resulting in the acquired $P_{low}$ and $P_{high}$. Finally, the low-pass and high-pass Fourier spectra are both adopted as the reconstruction target to better guide the model to capture local details and global information.}
    \label{fig_method}
    \vspace{-12pt}
\end{figure*}

An overview of the proposed SSL framework namely FreMIM is presented in Fig. \ref{fig_method}. 
Given an input medical image slice $X \in \mathbb{R}^{C \times H \times W}$ with a spatial resolution of $H \times W$ and $C$ channels (\# of modalities), the proposed foreground masking strategy is first employed on the original image to generate the masked image.
Then, the generic encoder (\ie according to various pre-training requirements, \textbf{both CNNs and Transformers encoder can be easily integrated into our framework}) takes the masked image as input, capturing the masked visual features through the hierarchical structure. 
After that, the encoded feature representations at different stages are jointly fed into our well-designed bilateral aggregation decoder, gradually producing the reconstructed Fourier spectrum with both low-level detail information and high-level semantic representation. 
By sequentially applying Fourier Transform on the original image and employing low/high-pass filters on the converted Fourier spectrum to acquire the expected reconstruction target, the reconstruction loss is applied to the similarity between the reconstructed spectrum and the expected low/high-pass spectrum target, realizing the helpful multi-stage supervision scheme with both low- and high-level representations in an end-to-end manner.

\vspace{-10pt}
\paragraph{Masking Strategy.}
As experimentally illustrated in several previous works\cite{bao2021beit,peng2022beit, xie2022simmim, he2022masked, gao2022convmae, tao2022siamese}, random masking strategy is not only simple but also effective for MIM-based self-supervised learning paradigm on large-scale natural images. However, different from natural images, the distribution of foreground and background pixels in medical images is extremely unbalanced. So randomly selecting spatial positions of a medical image would inevitably cause the generated mask to mostly cover background pixels and too many foreground pixels of the objects are reserved, counting against the model's reconstruction ability. To this end, we propose a simple yet effective foreground masking strategy to address this uneven distribution issue.

Specifically, given a binary mask $M \in \{0,1\}^{H \times W}$ initialized with zeros, its value at each spatial position is determined by whether the corresponding pixel value belongs to the foreground or not. If a pixel belongs to the foreground area, it will be filtered as one of the candidates to be masked during self-supervised pre-training. 
Since a medical image commonly consists of diverse channels, each one emphasizing a different foreground area, we take their overlapping parts as the final masked regions. The overall foreground masking strategy can be defined as:
\vspace{-5pt}
\begin{equation}
    \label{eq4}
    M_n(x,y)=\left\{
    \begin{array}{lr}
        0, & P_n(x,y) = 0 \\
        1, & P_n(x,y) \neq 0
    \end{array},
    \right.
\end{equation}
\vspace{-10pt}
\begin{equation}
    \mathcal{M} = M_1\cap M_2\cap M_3...\cap M_n,
\end{equation}
\vspace{-10pt}
\begin{equation}
    X_\mathcal{M} = \mathcal{M} \odot X,
\end{equation}
where $\odot$ is the Hadamard product, $P_n(x,y)$ represents the specific pixel value of the corresponding position $(x,y)$, $M_n$ denotes the generated mask of the specific image modality $M_n$. $\mathcal{M}$ and $X_\mathcal{M}$ respectively indicate the final mask of the original image and the masked image that will be fed into the model for the following reconstruction task. 

\vspace{-15pt}
\paragraph{Generic Encoder.}
As for the selection of encoder in our framework, FreMIM is not restricted to any specific kind of structure thanks to our pixel-wise foreground masking strategy.
Dislike some previous MIM-based methods can only be incorporated with various Vision Transformers (\eg Due to the random masking strategy of embedded image patches, MAE is only applicable for ViT\cite{vit} without the consideration of CNNs or hierarchical Transformer architecture), our FreMIM is a generic and flexible framework, which means both CNN-based and Transformer-based models can be easily integrated with our FreMIM for effective self-supervised pre-training.
Taking the aforementioned masked image as input, the network encoder gradually encodes the masked image slice with the hierarchical structure, producing the feature representations with diverse levels (\ie from low-level detail information to high-level semantics).
In this paper, three previous SOTA methods for medical image segmentation, \ie the representatives of the CNN-Transformer hybrid architectures and Vision Transformers, are selected as the backbones to validate the effectiveness of our method (more details are in Sec. \ref{experiments}).

\vspace{-12pt}
\paragraph{Multi-stage Supervision Scheme.}

Both low-level details and high-level global semantics are crucial, especially for medical image segmentation. The expectation of an effective SSL paradigm is to guide the visual backbone to learn the required representations with different levels through the hierarchical structure.
Following this intuition, we propose to design a multi-stage supervision scheme to fully supervise the representation learning of hierarchical stages.

As emphasized in Sec. \ref{sec:intro}, high-level and low-level information of an image is distributed in different frequency bands of the Fourier spectrum. So we propose to separately take advantage of the low-pass and high-pass Fourier spectrum as the supervision signal (\ie \textbf{reconstruction target}). One of the most intuitive ways is to utilize the identical high-pass Fourier spectrum to directly supervise multiple low-level stages and vice versa for low-pass counterparts.
However, there are mainly two drawbacks for this intuitive manner. 
On the one hand, this manner is kind of unreasonable and it violates the original intention of model learning at various low-level stages cause the feature representations learned at different low-level stages should be naturally different instead of the same. 
On the other hand, such a supervision method is too direct and simple, and does not make full use of
the correlation between the captured multi-stage features by the hierarchical structure to help the model better perform the MIM pretext task.

With regard to this, a well-designed \textbf{bilateral aggregation decoder} is proposed to better solve the reconstruction task in the frequency domain, further helping the encoder to learn a more generalized and more meaningful feature representation. 
Specifically, inside the proposed bilateral aggregation decoder, the encoded features at different stages are converged to the lowest stage (\ie with maximum spatial resolution) and the highest stage (\ie with minimum spatial resolution) in a bottom-up and top-down manner, respectively. 
In other words, the BAD separately aggregates the feature maps of different stages into the
lowest and highest resolution. 
Specifically, for ViT, the feature maps of layers 4th, 8th, and 12th are upsampled by 8, 4, and 2 times respectively to be fed to the BAD, following the deconvolution module in UNETR.
To be clear, 
the captured features of each adjacent stage will be fed into the convolutional block to achieve the strict alignment of both spatial resolution and channel dimension, which can be expressed as:
\begin{equation}	
    \hspace{-2.5mm} \!\mathbf{A}_{\text{low}}=\mathbf{Cat}(\mathbf{C}(S_{1}),\mathbf{Dc}(...,\mathbf{Cat}(\mathbf{C}(S_{n-1}),\mathbf{Dc}(S_{n})))\!, 
\end{equation}	
\vspace{-6mm}
\begin{equation}                            
    \hspace{-2.5mm}\!\mathbf{A}_{\text{high}}=\mathbf{Cat}(\mathbf{C}(S_{n}),\mathbf{Dc}(...,\mathbf{Cat}(\mathbf{C}(S_{2}),\mathbf{Dc}(S_{1})))\!, 
\end{equation}
where $\mathbf{A}_{\text{high}}$ and $\mathbf{A}_{\text{low}}$ separately denote the bilaterally aggregated high-level and low-level feature representations, $\mathbf{C}$, $\mathbf{Dc}$ and $\mathbf{Cat}$ indicate the convolutional block, deconvolution block, and concatenation operation respectively, $S_{i}$ denotes the feature maps output by the stage $i$.

Then, the aggregated feature representations at the lowest stage and highest stage will be mapped to the frequency domain through the introduced frequency mapping block (as illustrated in Fig. \ref{fig_method}), 
which are followed by the low-pass and high-pass filters to get the corresponding high-pass and low-pass prediction spectrum for the employed reconstruction loss. 
Specifically, the frequency mapping block (FMB) consists of a 2D-DFT, a Frequency Domain Perceptron (FDP), and a 2D-IDFT, which can be calculated as:
\begin{equation}
    \mathbf{P}_{\text{low}}=IDFT(W \odot DFT(\mathbf{A}_{\text{low}}) + b), 
\end{equation}
\vspace{-6mm}
\begin{equation}
    \mathbf{P}_{\text{high}}=IDFT(W \odot DFT(\mathbf{A}_{\text{high}}) + b), 
\end{equation}
where $DFT$ and $IDFT$ represent the Fast Fourier Transform and Inverse Fast Fourier Transform. W and b are both learnable parameters, $\odot$ is the Hadamard product.
In this way, a powerful SSL framework for \textbf{cross-domain reconstruction} is built with the benefit of the Fourier Transform's unique characteristics.
Although such a cross-domain reconstruction task is more difficult than intra-domain reconstruction, it can also assist the model in learning more robust feature representation, which is fully demonstrated in the following experimental section.

\subsection{Pre-training Strategy} 

\paragraph{Frequency Loss.} 
To alleviate the weight imbalance between different frequency band spectrums and facilitate the reconstruction of difficult frequency bands, we adopt focal frequency loss~\cite{jiang2021focal} as the loss function $\mathcal{L}_{\text{freq}}$ to implement gradient updating of weights for both low and high-frequency mapping, which is defined as:
\begin{align}
\tiny
\hspace{-2.8mm}
\!\mathcal{L}_{\text{freq}} =\frac{1}{H W} \sum_{u=0}^{H-1} \sum_{v=0}^{W-1} \omega(u, v)\odot \gamma(f(u, v),\hat{f}(u, v))^2\!, 
\end{align}	
where $f(u,v)$ is the predicted 2D-DFT of spatial frequency coordinate $(u,v)$ while $\hat{f}(u,v)$ denotes its corresponding Ground Truth value. $\gamma(f,\hat{f})$ calculates the squared Euclidean distance between actual and predicted values as their frequency distance. And $\omega$ is the spectrum weight matrix of a given location, which suppresses weights of easy frequencies. The calculation formulas are as follows:
\begin{align}
	\small
	&\omega(u, v)=\gamma(f(u, v),\hat{f}(u, v))^{\beta},\\
	&\gamma(f,\hat{f}) = \sqrt{(\mathcal{R}- \tilde{\mathcal{R}})^2+(\mathcal{I}- \tilde{\mathcal{I}})^2 },
\end{align}
where $\beta$ is a scaling factor for flexibility ($\beta$=1 by default) .   

\vspace{-5pt}
\paragraph{Overall Loss.}During pre-training, our FreMIM learns representation by solving content gestalt from both high-pass  and low-pass frequency:
\begin{align}
	\small
        \mathcal{L} = \mathcal{L}_{\text{freq}}(\mathbf{F}_{H}(\mathbf{P}_{\text{low}}), \mathbf{F}_{H}(\mathbf{T}))\\\notag
        + \alpha \mathcal{L}_{\text{freq}}(\mathbf{F}_{L}(\mathbf{P}_{\text{high}}), \mathbf{F}_{L}(\mathbf{T})),
	\label{overall-loss}
\end{align}
where $\mathbf{F}_{H}$ and $\mathbf{F}_{L}$ represent high-pass and 
low-pass frequency filter respectively. $\mathbf{T}$ indicates the original images. As shown in Fig. \ref{fig_method}, $\mathbf{P}_{\text{low}}$ is obtained by highest-stage and $\mathbf{P}_{\text{high}}$ is the opposite. $\alpha$ is the weight of high-level semantic information branches ($\alpha$ = 3 by default).

\begin{table*}
\centering
\caption{Comparison with previous self-supervised learning frameworks. `-' represents training from scratch.
Without introducing any extra samples, our FreMIM can consistently boost the model performance by a large margin compared with randomly initialized baselines.
}
\vspace{-3mm}
\label{tab:ablation4}
	\footnotesize
	\setlength{\tabcolsep}{1.2mm}
	\begin{tabular}{l l l|cccc}
		\toprule
  \multirow{2}{*}{Baseline} & \multirow{2}{*}{Backbone} & \multirow{2}{*}{Pre-train Method} & \multicolumn{4}{c}{Dice Score (\%) $\uparrow$} \\
        \cline{4-7}
		 & & & ET & WT & TC & Average\\
		\midrule
		TransBTSV2 \cite{li2022transbtsv2} & CNN-Transformer & - & 77.11 & 90.32 & 82.90 & 83.44\\
            \rowcolor{gray!18} TransBTSV2 \cite{li2022transbtsv2} & CNN-Transformer & FreMIM & \textbf{79.65} (\textcolor{red}{\textbf{+2.54}}) & \textbf{90.80} (\textbf{\textcolor{red}{+0.48}}) & \textbf{83.33} (\textbf{\textcolor{red}{+0.43}}) & \textbf{84.59} (\textbf{\textcolor{red}{+1.15}})\\
            \midrule
            UNETR \cite{hatamizadeh2022unetr} & ViT-B/16 \cite{dosovitskiy2020image} & - & 75.28 & 88.42 & 76.33 & 80.01\\
            UNETR \cite{hatamizadeh2022unetr} & ViT-B/16 \cite{dosovitskiy2020image} & MAE \cite{he2022masked} & 75.18 (\textcolor{blue}{-0.10}) & \textbf{88.95} (\textbf{\textcolor{red}{+0.53}}) & 78.47 (\textcolor{red}{+2.14}) & 80.87 (\textcolor{red}{+0.86})\\
            UNETR \cite{hatamizadeh2022unetr} & ViT-B/16 \cite{dosovitskiy2020image} & DINO \cite{caron2021emerging} & 75.22 (\textcolor{blue}{-0.06}) & 88.33 (\textcolor{blue}{-0.09}) & 75.89 (\textcolor{blue}{-0.44}) & 79.81 (\textcolor{blue}{-0.20})\\
            \rowcolor{gray!18} UNETR \cite{hatamizadeh2022unetr} & ViT-B/16 \cite{dosovitskiy2020image} & FreMIM & \textbf{76.50} (\textcolor{red}{\textbf{+1.22}}) & 88.86 (\textcolor{red}{+0.44}) & \textbf{78.82} (\textcolor{red}{\textbf{+2.49}}) & \textbf{81.39} (\textbf{\textcolor{red}{+1.38}})\\
		\midrule
            Swin UNETR \cite{tang2022self} & Swin-B \cite{liu2021swin} & - & 76.68 & 89.89 & 79.98 & 82.18\\
            Swin UNETR \cite{tang2022self} & Swin-B \cite{liu2021swin} & SimMIM \cite{xie2022simmim} & 77.59 (\textcolor{red}{+0.91}) & \textbf{90.47} (\textbf{\textcolor{red}{+0.58}}) & 80.34 (\textcolor{red}{+0.36}) & 82.80 (\textcolor{red}{+0.62})\\
            Swin UNETR \cite{tang2022self} & Swin-B \cite{liu2021swin} & Swin UNETR \cite{tang2022self} & 77.85 (\textcolor{red}{+1.17}) & 89.63 (\textcolor{blue}{-0.26}) & 78.65 (\textcolor{blue}{-1.33}) & 82.04 (\textcolor{blue}{-0.14})\\
            \rowcolor{gray!18} Swin UNETR \cite{tang2022self} & Swin-B \cite{liu2021swin} & FreMIM & \textbf{78.38} (\textcolor{red}{\textbf{+1.70}}) & 90.06 (\textcolor{red}{+0.17}) & \textbf{81.05} (\textcolor{red}{\textbf{+1.07}}) & \textbf{83.16} (\textbf{\textcolor{red}{+0.98}})\\
		\bottomrule
	\end{tabular}
 \vspace{-3mm}
\end{table*}

\section{Experiments and Results}
\label{experiments}

In this section, \textit{focusing on solely exploiting the given training samples (\ie the pre-training data only includes the specific downstream datasets without introducing any extra data) for 2D medical image segmentation (\eg solely BraTS 2019 is used for pre-training when evaluating brain tumor segmentation),} extensive experiments on three benchmark datasets are conducted to fully verify the effectiveness of FreMIM.  
{Note that the numbers between parenthesis represent the gains with respect to specific baselines trained from scratch, while the \textcolor{red}{red} and \textcolor{blue}{blue} color denote accuracy increase and decrease respectively.}
To save space, the visual comparisons in terms of segmentation and reconstruction results are presented in \textcolor{black}{appendix}.

\subsection{Experimental Setup}
\label{experimentalsetup}

\paragraph{Data and Evaluation Metrics.}
Our proposed method is evaluated on three benchmark datasets (\ie BraTS 2019 \cite{menze2014multimodal,bakas2017advancing,bakas2018identifying}, ISIC 2018 \cite{tschandl2018ham10000, codella2019skin} and ACDC 2017 \cite{bernard2018deep}) for medical segmentation. 
Due to space limit, more detailed elaborations are presented in the \textcolor{black}{appendix}.

\vspace{-15pt}
\paragraph{Implementation Details.}
The specific implementation details can be found in the \textcolor{black}{appendix}.

\subsection{Results and Analysis}

\paragraph{Comparison with Previous SSL Frameworks.} Based on five-fold cross-validation on the BraTS 2019 training set, we perform a fair comparison between our proposed FreMIM and previous self-supervised learning methods on various baselines including TransBTSV2 \cite{li2022transbtsv2}, UNETR \cite{hatamizadeh2022unetr}, and Swin UNETR \cite{tang2022self}, demonstrating the effectiveness and generalization capability of our FreMIM. 
For comprehensive comparisons, we select multiple self-supervised learning methods (\ie MAE \cite{he2022masked}, SimMIM \cite{xie2022simmim}, DINO \cite{caron2021emerging} and Swin UNETR \cite{tang2022self}), among which MAE and SimMIM have achieved promising results on natural images, DINO is a representative contrastive learning method, and Swin UNETR is a representative of the previous efforts on SSL methods for medical image analysis. 
\textit{To be clear, since some of these previous methods are limited to backbone structures (\eg MAE cannot be adapted to Swin Transformer backbone due to the token-dropping operation), for other methods we kept their original backbone as in their papers to achieve a fair comparison, which implicitly demonstrates our method's superior versatility to various backbones.}

As presented in Table \ref{tab:ablation4}, our FreMIM shows great superiority over all three baselines. 
Compared to training from scratch, the Average Dice scores on three baselines are simultaneously increased by 1.14\%, 1.38\%, and 0.98\% respectively after pre-training with our framework.
In comparison with MAE on UNETR and SimMIM on Swin UNETR, our FreMIM greatly improves model performance with the benefit of exploiting MIM in the frequency domain for global representation learning.
Since contrastive learning methods mainly focus on learning high-level semantics by instance discrimination task, neglecting the fine-grained representation learning results in poor results for UNETR with DINO pre-training.
In contrast, FreMIM takes advantage of the smooth structure information of organs and detailed contours and textures as supervision signals, better guiding the model's high-level and low-level representation learning.
Additionally, the Swin UNETR pre-training method achieves inferior performance. 
We believe the reasonable explanation for this phenomenon is that the Swin UNETR pre-training method heavily relies on the number of training samples to acquire useful prior knowledge (\ie it can not help models to capture the helpful representations as expected under the circumstance of limited pre-training samples).
On the contrary, without introducing any extra samples, our FreMIM can greatly boost model performance compared with random initialization, suggesting the effectiveness and data-efficient characteristic of our method.
In summary, our FreMIM with the advantages of the frequency domain is a generic and powerful MIM-based framework, which could bring consistent improvement in model performance without introducing extra data.

\vspace{-10pt}
\paragraph{Evaluation on Brain Tumor Segmentation.} Comparative experiments are also conducted on the BraTS 2019 validation set. As shown in Table \ref{table_sota} (a), our FreMIM achieves superior performance than previous methods with the competitive Dice scores of $79.74\%$, $90.23\%$, and $81.25\%$ on ET, WT, and TC respectively. In addition, it is notable that our method realizes a considerable decrease of Hausdorff distance on TC, reaching $6.934mm$. 
Without introducing any extra training samples, the proposed FreMIM can greatly boost model performance and outperform other previous SOTA approaches.
The considerable improvements made by FreMIM are powerful evidence of the effectiveness of using our method on MRI benchmarks.

\begin{table}[t]
    \caption{Performance comparisons on BraTS 2019, ISIC 2018 and ACDC 2017 datasets. Here TransBTSV2 denotes the 2D version of the original model to fit our proposed SSL framework.}
\vspace{-3mm}
\label{table_sota}
\footnotesize
\addtolength{\tabcolsep}{-2.5pt}
\begin{tabular}{l|c c c|c c c}
        \toprule[1.0pt]
        \multicolumn{7}{c}{\textbf{(a) BraTS 2019}}  \\ 
        \multirow{2}{*}{Method} & \multicolumn{3}{c|}{Dice Score (\%) $\uparrow$} & \multicolumn{3}{c}{Hausdorff Dist. (mm) $\downarrow$}\\
        \cline{2-7}
        &  ET &  WT &  TC &  ET &  WT & TC \\
        \hline
        \centering
        3D U-Net \cite{3dunet}  & 70.86 & 87.38 & 72.48 & 5.062 & 9.432 & 8.719 \\
        V-Net  \cite{vnet}    & 73.89 & 88.73 & 76.56 & 6.131 & 6.256 & 8.705 \\
        Attention U-Net  \cite{oktay2018attention}   & 75.96 & 88.81 & 77.20 & 5.202 & 7.756 & 8.258 \\
        Chen et al.  \cite{chen2019aggregating}   & 74.16 & \textbf{90.26} & 79.25 & 4.575 & \textbf{4.378} & 7.954 \\
        Li et al. \cite{li2019multi}           & 77.10 & 88.60 & 81.30 & 6.033 & 6.232 & 7.409 \\
        Frey et al. \cite{frey2019memory}      & 78.70 & 89.60 & 80.00 & 6.005 & 8.171 & 8.241 \\
        TransBTS  \cite{wang2021transbts}   & 78.36 &  88.89 & \textbf{81.41} & 5.908 & 7.599 & 7.584 \\
        TransUNet  \cite{chen2021transunet}   & 78.17 & 89.48 & 78.91 & 4.832 & 6.667 & 7.365 \\
        Swin-UNet  \cite{cao2021swin}   & 78.49 & 89.38 & 78.75 & 6.925 & 7.505 & 9.260 \\
        \hline
        TransBTSV2 \cite{li2022transbtsv2} & 78.63 & 90.09 & 80.23 & 3.729 & 6.194  & 7.725 \\
        \rowcolor{gray!18} \bf{TransBTSV2} & \textbf{79.74} & 90.23 & 81.25 & \textbf{3.209} & 5.875& \textbf{6.934} \\
        \rowcolor{gray!18} \bf{+FreMIM} & \textbf{\textcolor{red}{+1.11}} & \textbf{\textcolor{red}{+0.14}} & \textbf{\textcolor{red}{+1.02}} & \textbf{\textcolor{red}{-0.520}} & \textbf{\textcolor{red}{-0.319}} & \textbf{\textcolor{red}{-0.791}} \\ 
\end{tabular}
\addtolength{\tabcolsep}{1.5pt}
    \begin{tabular}{l|c c c c c}
        \toprule[1.0pt]
        \multicolumn{6}{c}{\textbf{(b) ISIC 2018}}  \\
        Method &  JI &  Dice &  Accuracy &  Recall &  Precision  \\
        \hline
        \centering
        U-Net \cite{ronneberger2015u} & 81.69 & 88.81 & 95.68 & 88.58 & 91.31  \\
        U-Net++ \cite{unet++} & 81.87 & 88.93 & 95.68 & 89.10 & 90.98  \\
        AttU-Net \cite{oktay2018attention} & 81.99 & 89.03 & 95.77 & 88.98 & 91.26  \\
        DeepLabv3+ \cite{chen2018encoder} & 82.32 & 89.26 & 95.87 & 89.74 & 90.87  \\
        CPF-Net \cite{feng2020cpfnet} & 82.92 & 89.63 & 96.02 & \textbf{90.62} & 90.71  \\
        BCDU-Net \cite{azad2019bi} & 80.84 & 88.33 & 95.48 & 89.12 & 89.68  \\
        Ms RED \cite{dai2022ms} & 83.45 & 89.99 & 96.19 & 90.49 & 91.47  \\
        \hline
        TransBTSV2 \cite{li2022transbtsv2} & 81.96 & 92.56 & 95.88 & 90.21 & 90.78  \\
        \rowcolor{gray!18} \bf{TransBTSV2}    & \textbf{83.53} & \textbf{93.39} &\textbf{96.44} & 90.18 & \textbf{92.61}  \\
        \rowcolor{gray!18} \bf{+FreMIM}    & \textbf{\textcolor{red}{+1.57}} & \textbf{\textcolor{red}{+0.83}} &\textbf{\textcolor{red}{+0.56}} & \textcolor{blue}{-0.03} & \textbf{\textcolor{red}{+1.83}}  \\ 
    \end{tabular}
\addtolength{\tabcolsep}{4.2pt}
    \begin{tabular}{l|c c c c }
        \toprule[1.0pt]
        \multicolumn{5}{c}{\textbf{(c) ACDC 2017}}  \\
        Method &  RV &  Myo &  LV &  Average  \\
        \hline
        \centering
        U-Net \cite{ronneberger2015u} & 86.91 & 87.17 & 90.65 & 88.25 \\
        AttU-Net \cite{oktay2018attention} & 86.78 & 86.93 & 91.84 & 88.52   \\
        Swin-UNet  \cite{cao2021swin} & 86.62 & 88.72 & 92.44 & 89.26   \\
        TransUNet  \cite{chen2021transunet} & 87.04 & 88.51 & \textbf{92.85} & 89.47   \\
        \hline
        TransBTSV2 \cite{li2022transbtsv2} & 86.80 & 87.76 & 91.87 & 88.81   \\
        \rowcolor{gray!18} \textbf{TransBTSV2} \cite{li2022transbtsv2} & \textbf{87.12} & \textbf{88.87} & 92.69 & \textbf{89.56} \\
        \rowcolor{gray!18} \bf{+FreMIM}  & \textcolor{red}{\textbf{+0.32}} & \textcolor{red}{\textbf{+1.11}} & \textcolor{red}{\textbf{+0.82}} & \textcolor{red}{\textbf{+0.75}} \\     
        \bottomrule[1.2pt]
    \end{tabular}
    \vspace{-5mm}
\end{table}

\vspace{-15pt}
\paragraph{Evaluation on Skin Lesion Segmentation.} We also verified the generality of FreMIM on RGB images dataset namely ISIC 2018 compared with the other seven well-performed algorithms. It could be seen from Table \ref{table_sota} (b) that, with the informative feature representations obtained from pre-training stages, our method could reach great performance on ISIC 2018 the five-fold cross-validation. Specifically, compared with previous SOTA methods, our results are higher on both JI, Dice, Accuracy, and Precision metrics. It is worth noting that our method promotes $\textbf{1.57\%}$ and $\textbf{1.83\%}$ on Dice score and Precision compared to training from scratch, demonstrating that FreMIM also presents strong capability on skin lesion segmentation.

\vspace{-15pt}
\paragraph{Evaluation on Cardiac Segmentation.} 
To evaluate the generalization ability of our proposed FreMIM, we also conduct experiments of cardiac segmentation on {MRI} scans utilizing the ACDC 2017 dataset \cite{bernard2018deep}. 
Since the official evaluation is supported by the online evaluation platform, the five-fold cross-validation is performed on ACDC 2017 training set.
The quantitative results on ACDC 2017 training set are presented in Table \ref{table_sota} (c). 
By guiding the baseline to better capture both the crucial high-level semantics and local detailed information, it is obvious that with boosted model performance in comparison with the baseline, our framework once again achieves comparable or even higher Dice scores than previous SOTA methods.

\subsection{Ablation Studies}
\label{ablationstudies}
\vspace{-2mm}
We conduct extensive experiments to prove the effectiveness of our FreMIM and validate its design rationale based on 5-fold cross-validation on BraTS 2019 training set, while TransBTSV2\cite{li2022transbtsv2} is selected as baseline for ablation study.

\vspace{-5pt}
\begin{table}
\centering
    \caption{Ablation study on the reconstruction target and supervision scheme.}
    \vspace{-3mm}
\label{tab:ablation1}
\footnotesize
\setlength{\tabcolsep}{1.1mm}
    \begin{tabular}{l l|cccc}
        \toprule
        \multirow{2}{*}{low-level target} & \multirow{2}{*}{high-level target} & \multicolumn{4}{c}{Dice Score (\%) $\uparrow$} \\
        \cline{3-6}
        &  & ET & WT & TC & Average\\
        \midrule
        - & - & 77.11 & 90.32 & 82.90 & 83.44\\
            high-pass & - & 77.82 & 90.60  & \textbf{83.60}  & 84.01(\textcolor{red}{+0.57})\\
            - & low-pass & 77.44  & 90.12  & 82.89  & 83.48(\textcolor{red}{+0.04})\\
            \midrule
          original image & original image & 79.33 & 90.23 & 81.95 & 83.83(\textcolor{red}{+0.39})\\
            all frequency & all frequency  & 79.12 & \textbf{90.80} & 82.58 & 84.17(\textcolor{red}{+0.73})\\
            low-pass & high-pass & 79.01 & 90.41 & 83.00 & 84.14(\textcolor{red}{+0.70})\\
            \rowcolor{gray!18} high-pass & low-pass & \textbf{79.65}  & \textbf{90.80}  & 83.33 & \textbf{84.59}(\textbf{\textcolor{red}{+1.15}})\\
        \bottomrule
    \end{tabular}
\vspace{-6mm}
\end{table}

\vspace{-10pt}
\paragraph{Reconstruction Target and Supervision Scheme.} Firstly, we explore the effect of different types of reconstruction targets and verify the effectiveness of our introduced multi-stage supervision scheme. The quantitative results are presented in Table \ref{tab:ablation1}. In comparison with random initialization in the first row, introducing either high-pass Fourier spectrum or low-pass counterpart as the reconstruction target at the corresponding low-level or high-level stage both lead to better segmentation performance to some extent. 
On the basis of this kind of single-level supervision manner, we further explore the effectiveness of a multi-level supervision scheme. 
As can be clearly seen in Table \ref{tab:ablation1} below the dividing line, simultaneously taking advantage of high-pass and low-pass frequency components, that carry abundant local details and global structural information, results in the best segmentation accuracy with the highest Average Dice Score of 84.59\%, fully demonstrating the powerful potential and rationale design of our FreMIM. 
No matter whether the reconstruction target is adjusted to the original image, the whole Fourier spectrum, or exchanged low/high-level target, it will all lead to a considerable decrease in model performance, which once again proves the strong theoretical rationale of exploiting FFT with the proposed FreMIM.

\vspace{-10pt}
\begin{table}[h]
	\centering
\caption{Ablation study on the masking strategy.}
\label{tab:ablation2}
\vspace{-3mm}
	\footnotesize
	\setlength{\tabcolsep}{2.5mm}
	\begin{tabular}{l | c  c  c  c  c}
		\toprule
          \multirow{2}{*}{Masking strategy} & \multicolumn{4}{c}{Dice Score (\%) $\uparrow$} \\
        \cline{2-5}
		 & ET & WT & TC & Average\\
		\midrule
		baseline & 77.11 & 90.32 & 82.90 & 83.44\\
            random mask & 79.07 & 90.64 & 83.19 & 84.30(\textcolor{red}{+0.86}) \\
            block wise mask & 79.03 & 90.00 & 82.11 & 83.71(\textcolor{red}{+0.27}) \\
            \rowcolor{gray!18} foreground mask & \textbf{79.65} & \textbf{90.80} & \textbf{83.33} & \textbf{84.59(\textcolor{red}{+1.15})}\\
		\bottomrule
	\end{tabular}
\vspace{-6mm}
\end{table}

\vspace{-3mm}
\paragraph{Masking Strategy.} Then we investigate the influence of different masking strategies to prove the effectiveness of the proposed foreground masking strategy. Table \ref{tab:ablation2} shows the performance comparison of our FreMIM with different masking strategies. It can be seen in Table \ref{tab:ablation2} that the original random masking leads to an accuracy increase $\uparrow$0.86\% on the Average Dice score, which is really promising.
However, by replacing vanilla random masking with our simple yet powerful foreground masking strategy, the model performance on segmentation tasks can be further boosted by a considerable margin, which shows the great superiority of selecting masked pixel candidates solely among foreground over conventional masking strategy.

\vspace{-10pt}
\begin{table}[h]
	\centering
\caption{Ablation study on the masking ratio.}
\label{tab:ablation3}
\vspace{-3mm}
	\footnotesize
	\setlength{\tabcolsep}{2.5mm}
	\begin{tabular}{c | c  c  c  c  c}
		\toprule
          \multirow{2}{*}{Masking Ratio} & \multicolumn{4}{c}{Dice Score (\%) $\uparrow$} \\
        \cline{2-5}
		 & ET & WT & TC & Average\\
		\midrule
		baseline & 77.11 & 90.32 & 82.90 & 83.44\\

            0.75 & 78.99 & 90.42 & 83.03 & 84.15(\textcolor{red}{+0.71}) \\
            0.50 & 79.19 & \textbf{90.80} & 83.18 & 84.39(\textcolor{red}{+0.95}) \\
            \rowcolor{gray!18} 0.25 & \textbf{79.65} & \textbf{90.80} & \textbf{83.33} & \textbf{84.59}(\textbf{\textcolor{red}{+1.15}}) \\
            0.15 & 79.37 & 90.23 & 82.88 & 84.16(\textcolor{red}{+0.72}) \\
            \midrule
            0.15, 0.20, 0.25 & 78.99 & 90.63 & \textbf{83.33} & 84.32(\textcolor{red}{+0.88}) \\
            0.25, 0.50, 0.75 & 79.23 & 90.62 & 82.88 & 84.24(\textcolor{red}{+0.80}) \\
		\bottomrule
	\end{tabular}
\vspace{-6mm}
\end{table}

\vspace{-3mm}
\paragraph{Masking Ratio.} After investigating the influence of various masking strategies, we further conduct experiments to seek the optimal masking ratio for our current framework.
As presented in Table \ref{tab:ablation3}, our FreMIM with a masking ratio of 0.25 achieves the best model performance. 
Once the masking ratio is either too low or too high, the reconstruction task in the frequency domain would be too easy or too hard, which may hinder the model from expected representation learning during self-supervised pre-training.
Besides, trying to take a step further, we also attempt to introduce a novel dynamic masking strategy (\ie the masking ratio gradually increases from the lowest to the highest during pre-training) for better guidance of the expected feature representation learning, which endows the SSL with easiest-to-hardest reconstruction level.
However, none of these attempts bring further accuracy improvements. 
Thus, the static masking strategy with a masking ratio of 0.25 is selected as our default setting.

\vspace{-3mm}
\paragraph{Choice of 2 Other Hyper-parameters.} 
Besides, we additionally conduct ablation studies on the loss weight $\alpha$ and the boundary definition (\ie frequency threshold) between high/low frequency, in Table \ref{tab:ablationonlossweight} and Table \ref{tab:frequencyboundary}, where PB denotes the specific value of frequency passband, showing the efficacy of our choice for these 2 hyper-parameters.

\begin{minipage}{0.48\textwidth}
\hspace{-5mm}
    \begin{minipage}[t]{0.48\textwidth}
		\makeatletter\def\@captype{table}
		\label{table1}
		\centering
    \caption{Ablation study on loss weight $\alpha$ during pre-training.}
    \label{tab:ablationonlossweight}
    \vspace{-3mm}
		\footnotesize
	\setlength{\tabcolsep}{1mm}
	\begin{tabular}{c | c  c  c  c  c}
		\toprule
          \multirow{2}{*}{$\alpha$} & \multicolumn{4}{c}{Dice Score (\%) $\uparrow$} \\
        \cline{2-5}
		 & ET & WT & TC & Average\\
           \hline
		0.5 & 79.34 & 90.11 & 82.16 & 83.87\\
            1 & 77.67 & 90.48 & 81.61 & 83.25 \\
            \rowcolor{gray!18} 3 & \textbf{79.65} & \textbf{90.80} & \textbf{83.33} & \textbf{84.59} \\
             5 & 78.93 & 90.64 & 82.98 & 84.18\\
		\bottomrule
	\end{tabular}
    \end{minipage}
    \hspace{1mm}
    \begin{minipage}[t]{0.48\textwidth}
		\makeatletter\def\@captype{table}
		\label{table2}
		\centering
\caption{Ablation study on high-/low-frequency boundary.}
\label{tab:frequencyboundary}
    \vspace{-3mm}
		\footnotesize
	\setlength{\tabcolsep}{1mm}
	\begin{tabular}{c | c  c  c  c  c}
		\toprule
          \multirow{2}{*}{PB} & \multicolumn{4}{c}{Dice Score (\%) $\uparrow$} \\
        \cline{2-5}
		 & ET & WT & TC & Average\\
           \hline
		  5 & 79.46 & 90.31 & 82.69 & 84.15\\
            \rowcolor{gray!18} 10 & \textbf{79.65} & \textbf{90.80} & \textbf{83.33} & \textbf{84.59} \\
            20 & 79.20 & 90.53 & 82.31 & 84.01 \\
            50 & 78.94 & 90.33 & 82.23 & 83.83\\
		\bottomrule
	\end{tabular}
\vspace{-5pt}
	\end{minipage}
\end{minipage}

\vspace{-9pt}
\begin{table}[htbp]
	\centering
\caption{Ablation study on the number of samples for self-supervised pre-training.}
\label{tab:ablationsupplementary}
\vspace{-3mm}
	\footnotesize
	\setlength{\tabcolsep}{2.2mm}
	\begin{tabular}{l | c  c  c  c  c}
		\toprule
          \multirow{2}{*}{Training samples} & \multicolumn{4}{c}{Dice Score (\%) $\uparrow$} \\
        \cline{2-5}
		 & ET & WT & TC & Average\\
           \hline
		baseline & 77.11 & 90.32 & 82.90 & 83.44\\
            0.3\%(\ie 1 sample) & 79.05 & 90.60 & 82.51 & 84.05(\textcolor{red}{+0.61}) \\
            10\% & 79.06 & 90.41 & \textbf{83.43} & 84.30(\textcolor{red}{+0.86}) \\
            \rowcolor{gray!18} 100\% & \textbf{79.65} & \textbf{90.80} & 83.33 & \textbf{84.59}(\textbf{\textcolor{red}{+1.15}}) \\
		\bottomrule
	\end{tabular}
\vspace{-6mm}
\end{table}

\vspace{-3mm}
\paragraph{Number of Pre-training Samples.}
Specifically, we further investigate the effect of different percentages of training samples used for our proposed FreMIM.
The quantitative results are presented in Table \ref{tab:ablationsupplementary}. 
It is clear in Table \ref{tab:ablationsupplementary} that the model performance is consistently improved with more and more employed training samples for the proposed FreMIM.
Besides, it is also surprising that by solely introducing 1 sample for pre-training our FreMIM can boost the model performance by a large margin (\ie $\uparrow 0.61\%$ on Average Dice score) compared with the randomly initialized baseline, demonstrating that our method is a data-efficient self-supervised learning paradigm.

\section{Conclusion}
In this paper, we presented the first study on exploring the powerful potential of MIM with frequency domain on pre-training deep learning models for medical image segmentation tasks. 
We focus on 2D medical image segmentation and proposed a new framework FreMIM taking advantage of both the rich global information and local details in the Fourier spectrum.
Deviating from the conventional paradigm as previous MIM methods, realizing reconstruction in the frequency domain empowers the framework with stronger representation learning capability.
Besides, by fully exploiting the specific characteristics contained in different frequency bands, the multi-stage supervision scheme can greatly boost the segmentation performance. 
Comprehensive experiments on three benchmark datasets quantitatively and qualitatively validated the effectiveness of our FreMIM, significantly improved the segmentation performance of baselines trained from scratch and showed superiority over state-of-the-art self-supervised approaches.

\vspace{-3mm}
\paragraph*{Acknowledgements}
Jianbo Jiao is supported by the Royal Society grant IES\textbackslash R3\textbackslash 223050.


{\small
\bibliographystyle{ieee_fullname}
\bibliography{egbib}
}

\appendix
\section*{Appendix}
In this appendix, we provide the following items:
\begin{enumerate}
    \item (Sec. \textcolor{red}{1}) More detailed information about the adopted three benchmark datasets (\ie BraTS 2019, ISIC 2018 and ACDC 2017).
    \item (Sec. \textcolor{red}{2}) Implementation details on the utilized three benchmark datasets (\ie BraTS 2019, ISIC 2018 and ACDC 2017).
    \item (Sec. \textcolor{red}{3}) More quantitative results about ablation studies of decoder structure and pre-training loss, as well as more experimental comparison on 3D baselines.
    \item (Sec. \textcolor{red}{4}) Visual comparison of reconstruction results and brain tumor segmentation results on BraTS 2019 dataset \cite{bakas2017advancing,bakas2018identifying, menze2014multimodal}, and skin lesion segmentation on ISIC 2018 dataset \cite{codella2019skin, tschandl2018ham10000} for qualitative analysis.
\end{enumerate}

\section{More Details about the Benchmark Datasets}

Our proposed method is evaluated on three benchmark datasets for medical segmentation.
The Brain Tumor Segmentation 2019 challenge (\textbf{BraTS 2019}) dataset \cite{menze2014multimodal,bakas2017advancing,bakas2018identifying} is composed of multi-institutional pre-operative MRI sequences, including 335 patient cases for training and 125 cases for validation. 
Each sample contains four modalities (FLAIR, T1, T1c, T2) with the size of $240\times240\times155$, and the corresponding ground truth consists of 4 classes: background (label 0), necrotic and non-enhancing tumor (label 1), peritumoral edema (label 2) and GD-enhancing tumor (label 4). 
The Dice score and the Hausdorff distance (95\%) metrics are used for evaluating the segmentation accuracy of different regions, including enhancing tumor region (ET, label 4), regions of the tumor core (TC, labels 1 and 4), and the whole tumor region (WT, labels 1,2 and 4).
The International Skin Imaging Collaboration 2018 (\textbf{ISIC 2018}) dataset \cite{tschandl2018ham10000, codella2019skin} is a collection of 2594 RGB images of skin lesion for training, around 100 samples for validation, and 1000 samples for testing. 
Five metrics are specifically employed for the quantitative assessment of model performance, including Dice, Jaccard Index (JI), Accuracy, Recall, and Precision.
The Automated Cardiac Diagnosis Challenge 2017 (\textbf{ACDC 2017}) dataset \cite{bernard2018deep} is collected from different patient cases using MRI scanners, including 3D cardiac MRI cine for both end-diastolic (ED) and end-systolic (ES) phases instances. The publicly available training dataset consists of 100 patient scans, which are split into 80 training samples and 20 testing samples. The ground truth contains 3 classes: right ventricle (RV), myocardium (Myo) and left ventricle (LV).

\section{Implementation Details}

\begin{table}[htbp]
\footnotesize
\addtolength{\tabcolsep}{10.6pt}
    \begin{tabular}{c|c c}
        \toprule[1.0pt]
        Config &  Pre-training &  Fine-tuning  \\
        \hline
        \centering
        optimizer & Adam & Adam \\
        base learning rate & $10^{-4}$ & $10^{-4}$  \\
        weight decay & $10^{-5}$ & $10^{-5}$  \\
        batch size  & 64 & 64 \\
        lr decay schedule & cosine decay & cosine decay \\
        training epochs  & 250 & 500 \\     
        \bottomrule[1.2pt]
    \end{tabular}
\vspace{3pt}
\caption{Training settings on BraTS 2019 dataset.}
\label{table_implementationBraTS2019}
\end{table}

\begin{table}[htbp]
\footnotesize
\addtolength{\tabcolsep}{10.6pt}
    \begin{tabular}{c|c c}
        \toprule[1.0pt]
        Config &  Pre-training &  Fine-tuning  \\
        \hline
        \centering
        optimizer & SGD & SGD \\
        base learning rate & $10^{-3}$ & $5\times10^{-4}$  \\
        weight decay & $10^{-8}$ & $10^{-8}$  \\
        batch size  & 12 & 12 \\
        lr decay schedule & poly & poly \\
        training epochs  & 125 & 300 \\     
        \bottomrule[1.2pt]
    \end{tabular}
\vspace{5pt}
\caption{Training settings on ISIC 2018 dataset.}
\label{table_implementationISIC2018}
\end{table}

\begin{table}[htbp]
\footnotesize
\addtolength{\tabcolsep}{10.6pt}
    \begin{tabular}{c|c c}
        \toprule[1.0pt]
        Config &  Pre-training &  Fine-tuning  \\
        \hline
        \centering
        optimizer & SGD & SGD \\
        base learning rate & $10^{-2}$ & $10^{-2}$  \\
        weight decay & $10^{-4}$ & $10^{-4}$  \\
        batch size  & 16 & 16 \\
        lr decay schedule & poly & poly \\
        training epochs  & 300 & 1200 \\     
        \bottomrule[1.2pt]
    \end{tabular}
\vspace{5pt}
\caption{Training settings on ACDC 2017 dataset.}
\label{table_implementationacdc2017}
\end{table}

The proposed method is implemented in PyTorch \cite{paszke2019pytorch} and trained with two NVIDIA Geforce RTX 3090 GPUs. 
The specific training hyper-parameter configurations of our FreMIM on BraTS 2019, ISIC 2018 and ACDC 2017 can be found in Table \ref{table_implementationBraTS2019}, \ref{table_implementationISIC2018}, \ref{table_implementationacdc2017} respectively.

\section{More Quantitative Results.}

\paragraph{Importance of the bilateral aggregation decoder (BAD) and focal loss:}
We also conduct supplementary ablation studies to validate the effectiveness of BAD and focal loss,
in Table \ref{tab:decoderandloss}, which clearly justifies the importance and effectiveness of our design choices.
\begin{table}[!h]
	\centering
	\footnotesize 
    \setlength{\tabcolsep}{2.2mm}
	\begin{tabular}{c c | c  c  c  c  c}
		\toprule
          \multirow{2}{*}{Decoder} & \multirow{2}{*}{Loss} & \multicolumn{4}{c}{Dice Score (\%) $\uparrow$} \\
        \cline{3-6}
		 & & ET & WT & TC & Average\\
           \hline
		  Single & Focal & 77.88 & 90.31 & 82.01 & 83.40\\
            BAD & L1 & 78.75 & 90.83 & 82.19 & 83.92(\textcolor{red}{+0.48}) \\
            BAD & MSE & 79.18 & 90.47 & 82.79 & 84.15(\textcolor{red}{+0.71}) \\
            \rowcolor{gray!18} BAD & Focal  & \textbf{79.65} & \textbf{90.80} & \textbf{83.33} & \textbf{84.59}(\textcolor{red}{+1.15}) \\
		\bottomrule
	\end{tabular}
\caption{Ablation study on the type of decoder and loss function for self-supervised pre-training.}
\label{tab:decoderandloss}
\vspace{-18pt}
\end{table}

\paragraph{Evaluations on 3D baselines:}
Noticeably, our framework is easily extendable to 3D version, enhancing 3D baseline's performance. 
To convince this point, we also conduct experiments on a commonly used 3D benchmark dataset BTCV \cite{landman2015miccai}, with 3D UNet and 3D Swin UNETR\cite{tang2022self} as 3D baselines for comparison. 
The employed pre-training methods (\ie Model genesis \cite{zhou2019models} and Swin UNETR \cite{tang2022self}) are both previous efforts on SSL for medical image analysis.
We follow the same pre-training and fine-tuning settings as in Swin UNETR for a fair comparison.
Besides, we evaluate the effectiveness of our approach in terms of five-fold cross-validation on the training set and the evaluation metrics stay the same as in Swin UNETR.
Results in Table \ref{3Dbaselines_and_3Ddatasets} provide substantial evidence of our method's generalization ability and potential.
\vspace{-10pt}
\begin{table}[!h]
	\centering
	\footnotesize
	\setlength{\tabcolsep}{0.5mm}
	\begin{tabular}{c | c  c  c  c  c}
		\toprule
            Method & Scratch & Models genesis\cite{zhou2019models} & Swin UNETR\cite{tang2022self} & Ours\\
		\midrule
		3D UNet  & 80.41 & 81.25 & - & \textbf{81.72}\\
        Improvement$\uparrow$ & - & (\textcolor{red}{+0.84}) & - &(\textcolor{red}{+1.31})\\
		\midrule
        Swin UNETR 3D  & 81.06 & - & 82.25 & \textbf{82.80} \\
        Improvement$\uparrow$ & - & - & (\textcolor{red}{+1.19}) &(\textcolor{red}{+1.74})\\
		\bottomrule
	\end{tabular}
\caption{Comparison with previous SSL works on BTCV dataset.}
\label{3Dbaselines_and_3Ddatasets}
\end{table}

\section{Visual Comparison for Qualitative Analysis}

\begin{figure*}[htbp]
    \centering
    \vspace{-5pt}
    \includegraphics[width=0.85\textwidth]{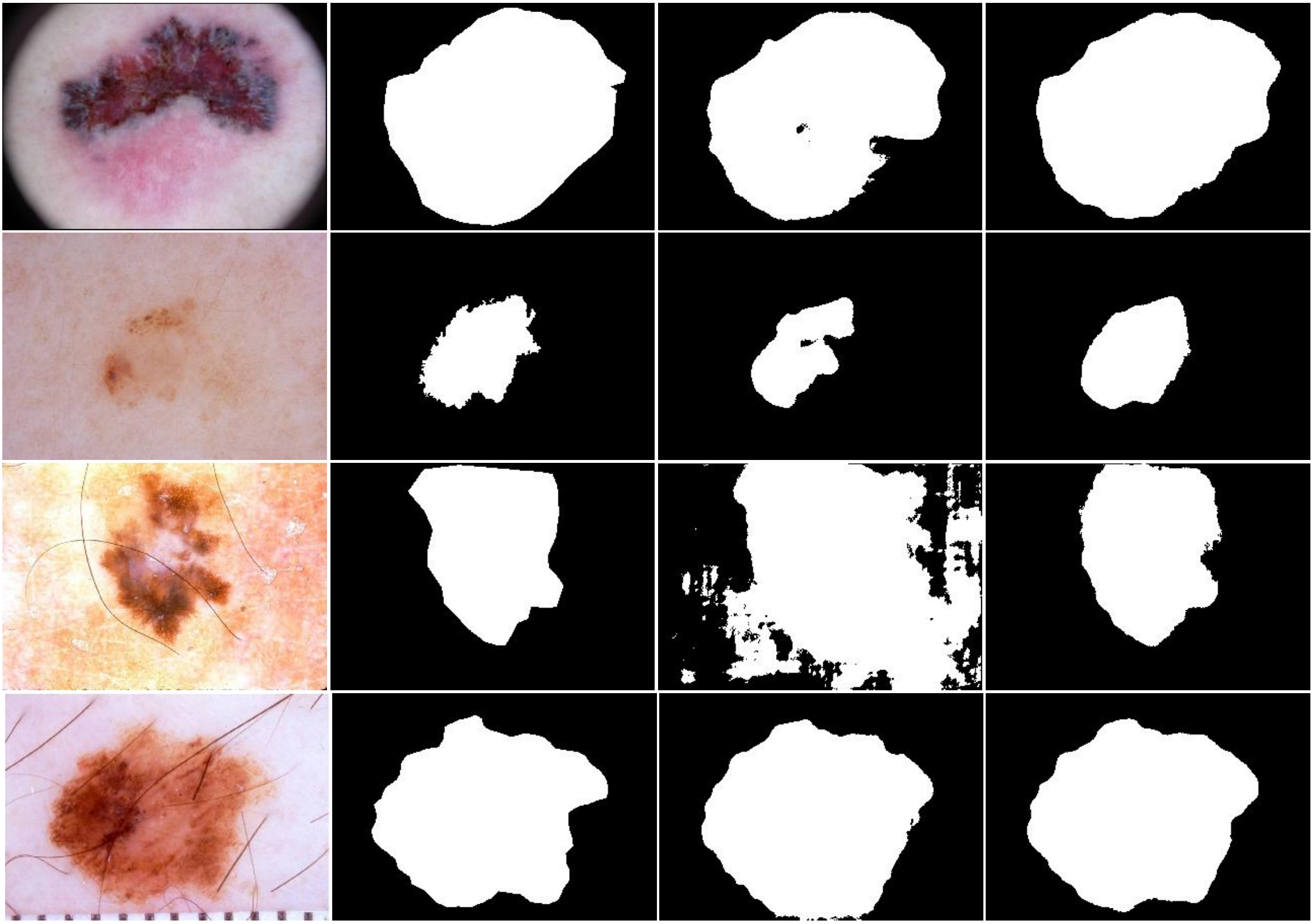}
    \begin{tabu} to 0.85\linewidth{X[1.0c] X[1.0c] X[1.0c]  X[1.0c]} 
        \scriptsize{Image} &  \scriptsize{Ground Truth} &  \scriptsize{Baseline} &  \scriptsize{Ours} \\
    \end{tabu}
    \caption{The visual comparison of skin lesion segmentation results on ISIC 2018 dataset with TransBTSV2 as the baseline.}
    \label{fig_supplementary_ISIC2018}
    \vspace{-5pt}
\end{figure*}

\begin{figure*}[htbp]
    \centering
    \includegraphics[width=0.85\textwidth]{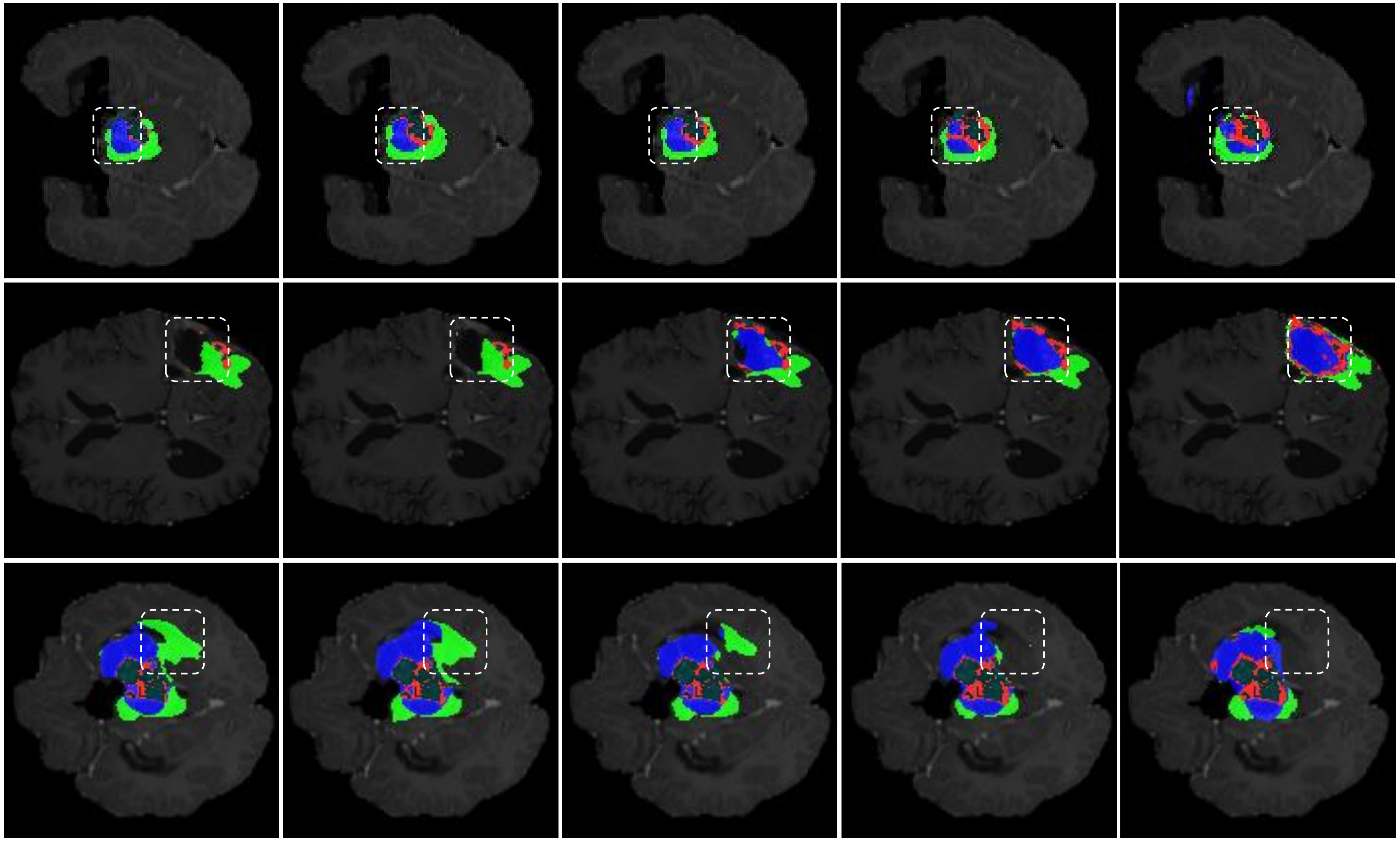}
    \begin{tabu} to 0.85\linewidth{X[1.0c] X[1.0c] X[1.0c] X[1.0c] X[1.0c]} 
        \scriptsize{Baseline} &  \scriptsize{DINO} &  \scriptsize{MAE} &  \scriptsize{\textbf{Ours}} &  \scriptsize{GT} \\
    \end{tabu}
    \caption{The visual comparison of MRI brain tumor segmentation results with UNETR as baseline. The \textcolor{blue}{blue} regions denote the enhancing tumors, the \textcolor{red}{red} regions denote the non-enhancing tumors and the \textcolor{green}{green} ones denote the peritumoral edema.
    }
    \label{fig3}
\end{figure*}

\noindent \textbf{Segmentation Results.} 
Firstly, the skin lesion segmentation results on ISIC 2018 dataset is presented in Fig. \ref{fig_supplementary_ISIC2018}. It can be obviously seen that the model can generate much more accurate and fine-grained segmentation masks compared with baseline with the benefit of employing our proposed FreMIM.
Simultaneously, we compare the segmentation performance of different self-supervised methods, including MAE, DINO, and FreMIM on the BraTS 2019 dataset with visualization results. As shown in Fig. \ref{fig3}, our method promotes the detailed pixel delineation of brain tumors and obtains more accurate predictions.

\begin{figure*}[htbp]
    \centering
    \includegraphics[width=0.95\textwidth]{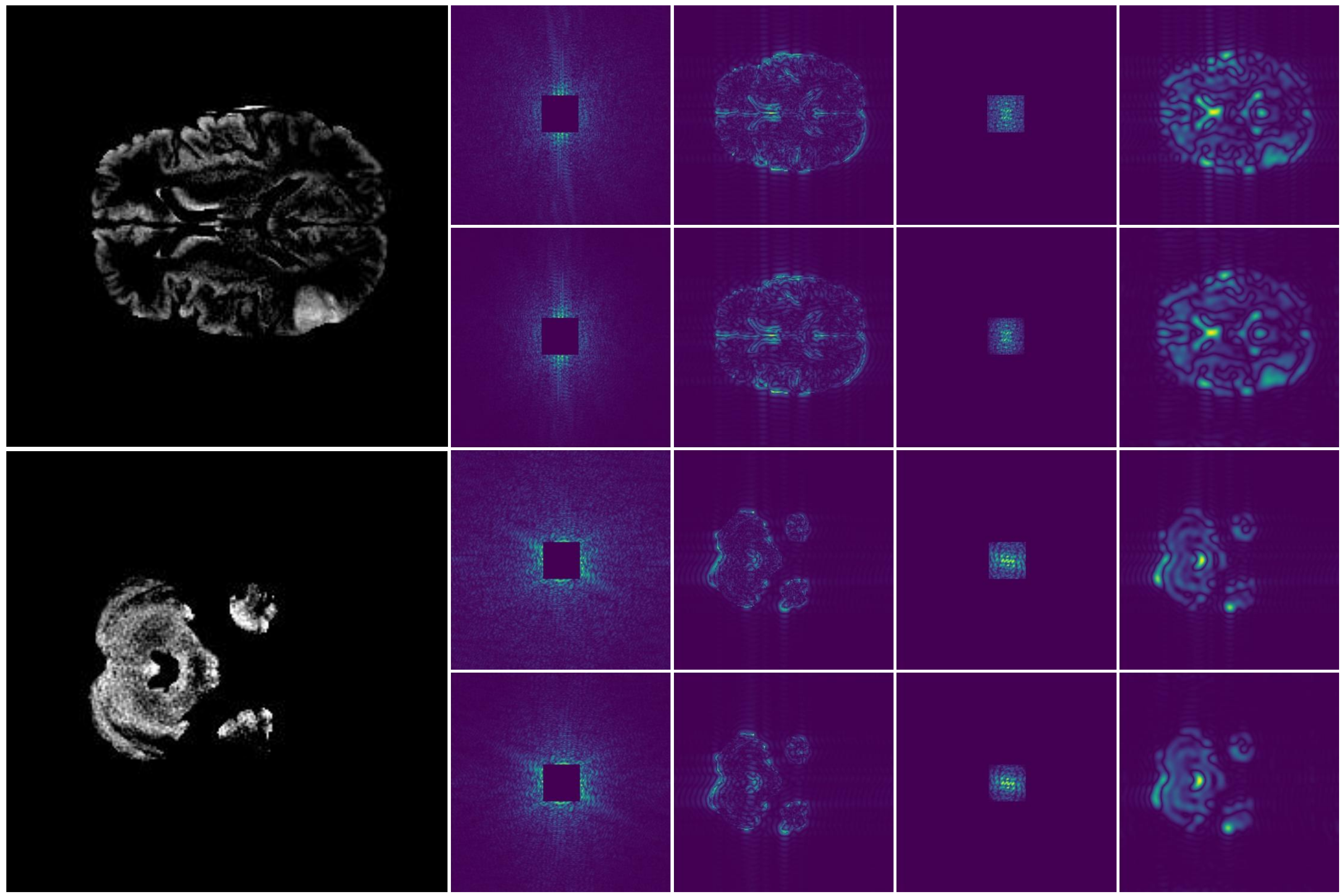}
    \rotatebox{90}
    {
    \begin{tabu} to 0.62\linewidth{X[1.0c] X[1.0c] X[1.0c] X[1.0c]} 
        \scriptsize{(b) Prediction} & \scriptsize{(a) Target} & \scriptsize{(b) Prediction} & \scriptsize{(a) Target}\\ 
    \end{tabu}
    }
    \begin{tabu} to 0.94\linewidth{X[1.78c] X[0.95c] X[0.9c] X[0.8c] X[1.0c]} 
        \scriptsize{Image} &  \scriptsize{High Frequency} &  \scriptsize{High Frequency-iFFT} &  \scriptsize{Low Frequency} &  \scriptsize{Low Frequency-iFFT} \\
    \end{tabu}
    \caption{The visualization of reconstruction results by our FreMIM in the frequency domain.}
    \label{fig_supplementary_brats2019}
\end{figure*}

\noindent \textbf{Reconstruction Results.} 
To convincingly prove the superiority of our FreMIM, we further supplement more visual comparison of reconstruction results on BraTS 2019 dataset for qualitative analysis.
As is shown in Fig. \ref{fig_supplementary_brats2019}, our method can nicely achieve the reconstruction task of Fourier spectrum and generate the corresponding reconstruction spectrum approximately the same as original image.
To be mentioned, for each image slice, the first row is the original image and the second row is our reconstruction results of the Fourier spectrum.

\end{document}